\pdfoutput=1
\newcommand{\DatasetName}{\textsc{WithdrarXiv}}

\documentclass[11pt]{article}

\usepackage[preprint]{acl}

\usepackage{times}
\usepackage{latexsym}
\usepackage{amsfonts}

\usepackage[T1]{fontenc}

\usepackage[utf8]{inputenc}

\usepackage{microtype}

\usepackage{inconsolata}

\usepackage{graphicx}
\usepackage{hyperref}
\usepackage{todonotes}
\usepackage{enumitem}
\usepackage{mdframed}

\title{\DatasetName: A Large-Scale Dataset for Retraction Study}

\author{Delip Rao\thanks{corresponding author}\\
  University of Pennsylvania\\
  \texttt{delip@seas.upenn.edu} \\\And
  Jonathan Young\\
  \url{arXiv.org} \\
  \texttt{jonathan@arxiv.org} \\\And
  Thomas Dietterich\\
  \hspace{-2mm}Oregon State University\\
  \texttt{tgd@cs.orst.edu} \\\And
  Chris Callison-Burch\\
  \hspace{3mm}University of Pennsylvania\\
  \texttt{ccb@seas.upenn.edu} \\}

\begin{document}
\maketitle

\begin{abstract}
Retractions play a vital role in maintaining scientific integrity, yet systematic studies of retractions in computer science and other STEM fields remain scarce. We present \DatasetName, the first large-scale dataset of withdrawn papers from arXiv, containing over 14,000 papers and their associated retraction comments spanning the repository's entire history through September 2024. Through careful analysis of author comments, we develop a comprehensive taxonomy of retraction reasons, identifying 10 distinct categories ranging from critical errors to policy violations. We demonstrate a simple yet highly accurate zero-shot automatic categorization of retraction reasons, achieving a weighted average F1-score of 0.9594. Additionally, we release\footnote{\url{https://github.com/darpa-scify/withdrarxiv}} \DatasetName-\textsc{SciFy}, an enriched version including scripts for parsed full-text PDFs, specifically designed to enable research in scientific feasibility studies, claim verification, and automated theorem proving. These findings provide valuable insights for improving scientific quality control and automated verification systems. Finally, and most importantly, we discuss ethical issues and take a number of steps to implement responsible data release while fostering open science in this area.
\end{abstract}

\begin{figure*}[t]
    \centering
    \includegraphics[width=1\linewidth]{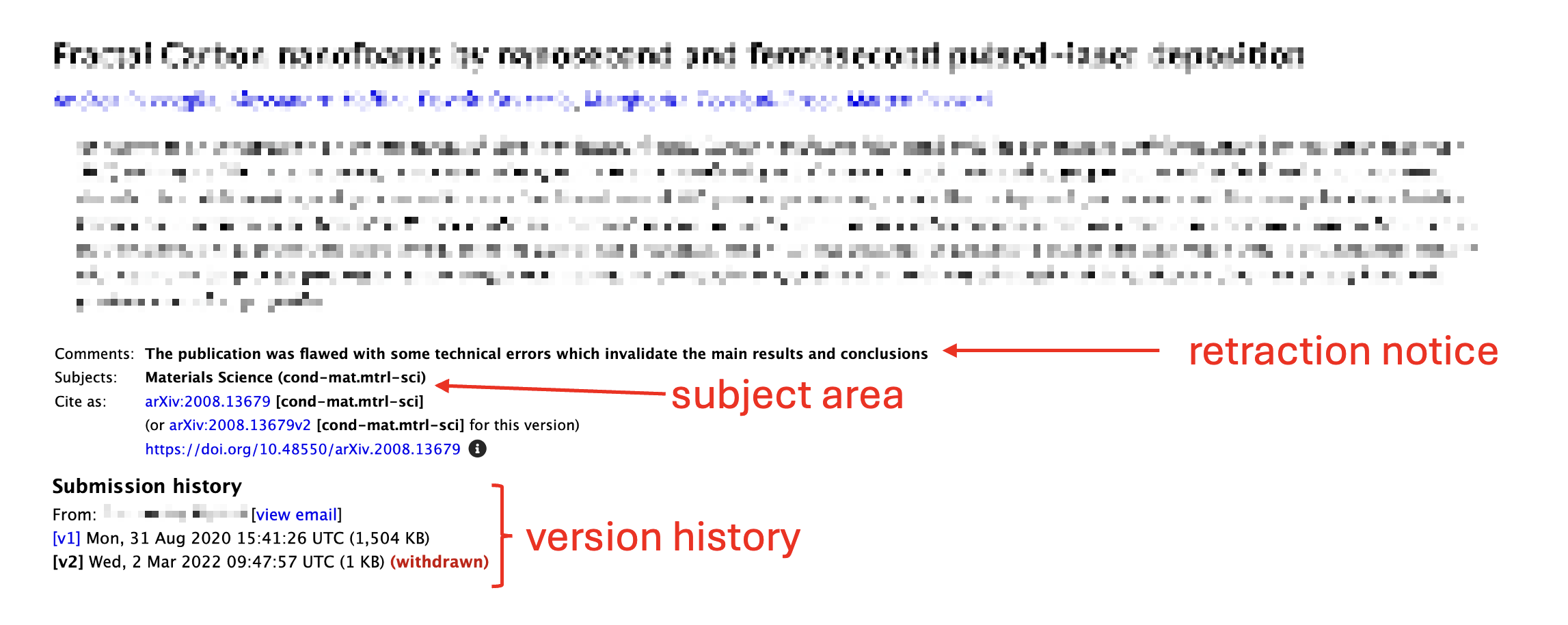}
    \caption{Metadata elements extracted from arXiv abstract pages for building \DatasetName}
    \label{fig:arxiv-meta-data}
\end{figure*}

\section{Introduction}
Retraction is an essential and ethical part of the scientific process, providing authors and publishers opportunities to alert readers to publications that contain serious flaws, erroneous data, or generally unreliable conclusions \cite{Katavi2014}. In certain academic communities, especially biomedical, organized discipline-specific retraction studies are common and deemed essential in maintaining the integrity of their respective scientific bodies. See (\citealp{Levett2023PublicationRI}; \citealp{Call2024ASR}; \citealp{Wang2017}), for example.

While regular retraction studies are common in medicine (see Section~\ref{sec:related-work} for examples), they are notably rare, or even absent, in Computer Science and other science/engineering fields. These fast-paced communities increasingly rely on preprint servers like \href{https://arXiv.org}{arXiv.org} to quickly disseminate research. However, to our knowledge, no systematic retraction studies have been conducted on arXiv preprints\footnote{Not considering one-off publicized retractions, such as the \href{https://en.wikipedia.org/wiki/LK-99}{LK-99 episode}.}. With generative AI-driven science gaining prominence (\citealp{Agarwal2024LitLLMAT}; \citealp{Kasanishi2023SciReviewGenAL}; \citealp{lu2024aiscientistfullyautomated}) and retrieval augmented systems increasingly depending on preprint servers for knowledge, it becomes crucial to understand withdrawn preprints and ensure their exclusion. In fact,~\citep{Pfeifer1990} conclude that ``a dearth of available information on retracted works; inconsistency in retraction format, terminology, and indexing'' as leading causes for retracted works to continue being cited post-retraction. Furthermore, studying retracted works offers opportunities to design and automate scientific feasibility\footnote{Is a proposed scientific method, idea, or technique feasible or reproducible?} techniques as we elaborate in Section~\ref{sec:enabling-scify}.

In this work, we make several key contributions to address these challenges. First, we introduce \DatasetName, the first comprehensive dataset of withdrawn papers from arXiv, containing over 14,000 withdrawn papers and their associated retraction comments spanning arXiv's entire history through September 2024. Second, we develop a taxonomy of retraction reasons by analyzing author comments, identifying 10 distinct categories that provide insights into why researchers withdraw their work. Third, we demonstrate the effectiveness of large language models in automatically categorizing retraction reasons, achieving a weighted average F1-score of 0.96 across all categories. Finally, we release \DatasetName-\textsc{SciFy}, an enriched version of a subset of our dataset that includes scripts for producing parsed full-text PDFs, specifically designed to enable research in scientific feasibility studies. Our work not only provides valuable insights into the patterns and reasons for scientific retractions but also creates resources that can help improve the integrity and efficiency of the scientific process.

\section{Why arXiv?}
ArXiv, the pioneering open-access repository for scientific pre/post-prints, has become an indispensable resource for scholars worldwide in STEM areas. Since its inception in 1991, arXiv has grown exponentially, now hosting over 2.5 million scholarly articles across various scientific disciplines. The platform's impact~\cite{arxiv2023} is evident in its staggering usage statistics: as of October 2023, arXiv had facilitated over 3 billion total downloads, with more than 5 million monthly active users. The repository's growth shows no signs of slowing, with over 2.2 million total submissions by the end of 2022, and this number increasing to approximately 2.6 million by November 2024\footnote{Derived from: \url{https://arxiv.org/stats/monthly_submissions}}.

Several features of arXiv have contributed to this exponential growth of the platform, including easy and rapid dissemination of scholarly work, an open-access model, and versioning capabilities, that enable authors to update their articles as their research progresses or in response to feedback. Additionally, the platform allows for the withdrawal of articles, giving researchers opportunities to maintain scientific integrity in the era of rapid publishing. The versioning and withdrawal features of arXiv provide a unique opportunity to study scientific retractions on a large scale.

Finally, our interest in arXiv is also because of the growing interest in the platform by developers of retrieval augmented scientific reasoning systems. 

\section{Building the \DatasetName~Dataset}
In this section, in the interest of reproducibility, we introduce the four steps of building the \DatasetName~dataset.
\subsection{Step 1: Harvesting Withdrawn ArXiv ids}
We worked with \href{arxiv.org}{arXiv.org} to collect all withdrawn article IDs on arXiv as of September 19, 2024. While this information is public, it is non-trivial to collect them without the support of arXiv or expending resources to filter the entire arXiv data dump. This effort produced a list of 16,460 article IDs. These are not distinct articles but sometimes include multiple revisions, as identified by their version numbers, of the same article. For example, for arXiv:2309.11721, versions 3 and 5 are marked as "withdrawn," and these are represented as two entries in our dataset -- arXiv:2309.11721v3 and arXiv:2309.11721v5 respectively. In this dataset, around 11\% of the arXiv identifiers represent different versions of a paper.

\subsection{Step 2: Comment Extraction}
\label{sec:comment-extraction}
Every arXiv identifier comes with a comment section, subject areas the paper belongs to, and a list of version URLs that allow us to backtrack to the full paper that was withdrawn. See Figure~\ref{fig:arxiv-meta-data}. We crawled \href{arxiv.org}{arXiv.org} abstract pages for the identifiers of interest and extracted these elements by parsing HTML pages. The crawl yielded a dataset of size 16,395. Before any further processing, we scrub the dataset of any personally identifiable information (PII). While PII is rare in arXiv withdrawal comments, we did spot a few names and email addresses. We used \texttt{scrubadubdub}, a Python package using NLP techniques~\cite{Kylemclarenscrub}, for replacing PII with placeholders like [RETRACTED\_NAME] and [RETRACTED\_EMAIL].

\subsection{Step 3: Comment Clustering}
\label{sec:comment-clustering}
The retraction comments are free-form natural language text and require categorization for careful analysis. To derive the comment categories, we first embed the comments using an off-the-shelf text embedding model~\cite{nussbaum2024} and then cluster these embeddings using K-means. To ensure we do not miss nuances, we generated a large number of clusters (K=100), and manually reviewed the clusters, identifying categories and hard test cases for each category. Our labeling produced the following 10 categories:\\

\begin{enumerate}[nolistsep]
    \item Factual/methodological/other critical errors in manuscript
    \item Incomplete exposition or more work in progress
    \item Typos in manuscript
    \item Self-identified as ``not novel''
    \item Administrative or legal issues
    \item ArXiv policy violation
    \item Subsumed by another publication
    \item Plagiarism
    \item Personal reasons
    \item Reason not specified
\end{enumerate}\leavevmode\newline
\noindent We provide detailed explanations for each of the categories with examples in Section~\ref{sec:retraction-categories}.

\subsection{Step 4: Zero-shot Comment Categorization}
\label{sec:zero-shot-comment-cat}
To map the comments to one of the 10 categories in Section~\ref{sec:comment-clustering}, we use the \texttt{gpt-4} model\footnote{accessed on Oct 10th, 2024} in a zero-shot setup. 
\begin{mdframed}[backgroundcolor=blue!20]
\textbf{Zero-shot prompt for Comment Categorization:}\\

\noindent You are given a comment from a paper withdrawal. Your task is to classify the comment into one of the following categories: ``incomplete exposition or more work in progress'', ``factual/methodological/other critical errors in manuscript'', ``typos in manuscript'', ``subsumed by another publication'', ``not novel'', ``plagiarism'', ``administrative or legal issues'', ``arXiv policy violation'', ``reason not specified'', ``personal reasons''. Return the category in a JSON format \{"category": <category>\}.
\end{mdframed}
We tested the prompt with the hard cases identified in Section~\ref{sec:comment-clustering} and verified that all identified hard cases passed. We include a full list of the test cases along with the accompanying code release; Table~\ref{tab:test-cases} gives a sample.

\begin{table}
    \centering
    \begin{tabular}{p{0.5\linewidth} | p{0.5\linewidth}}
          \hline
          "This paper has been withdrawn by the author. Please see arXiv:0806.0780" & Subsumed by another publication \\
          \hline
          "the data set did not pass the IRB review" & Administrative or legal issues\\
          \hline
          "60F15" (sic) & Reason not specified\\
          \hline
    \end{tabular}
    \caption{A sample of the test cases used during prompt creation}
    \label{tab:test-cases}
\end{table}

\section{Evaluation of Zero-shot categorization}
To evaluate how well the zero-shot prompting in Section~\ref{sec:zero-shot-comment-cat} performs, we manually annotated a subset of the 16K comments. We selected this subset using stratified sampling of each category, choosing 10\% of the comments in each category or 50 if the 10\% was less than 50, or all if the total in the category was less than 50. This resulted in 1,620 comments that were hand-labeled.
\begin{figure*}[h]
    \centering
    \includegraphics[width=1\linewidth]{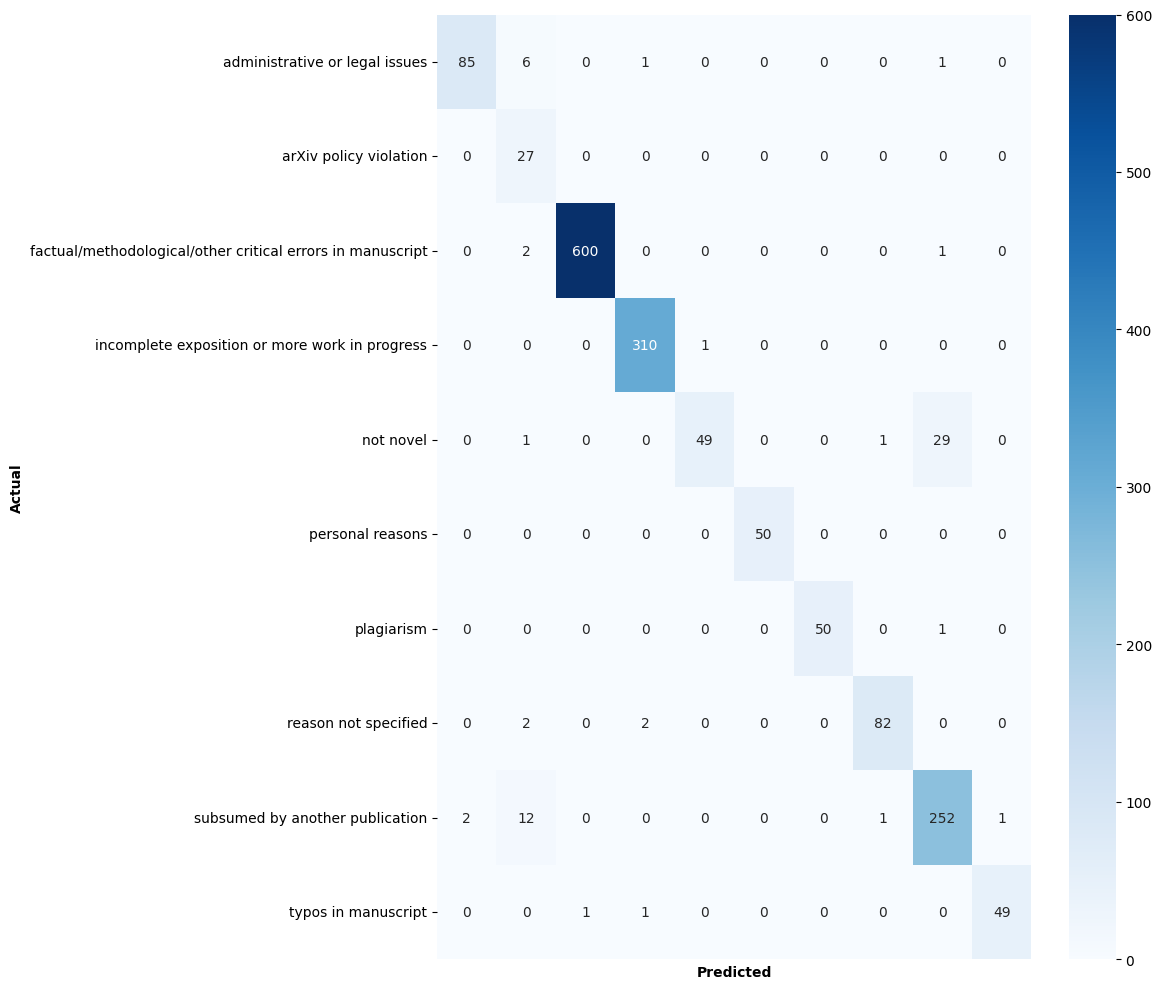}
    \caption{Confusion matrix for zero-shot categorization as evaluated on human annotations of a 10\% stratified sample of the comments (total support=1620)}
    \label{fig:zero-shot-cm}
\end{figure*}

\begin{table}[h]
\centering
\begin{tabular}{p{0.7\linewidth} | p{0.1\linewidth}}
\hline\hline
\textbf{Category label }& \textbf{F1-score }\\
\hline
Administrative or legal issues & 0.9444 \\
ArXiv policy violation & 0.7013 \\
Factual/methodological/other critical errors in manuscript & 0.9967 \\
Incomplete exposition or more work in progress & 0.9920 \\
Self-identified as ``not novel'' & 0.7538 \\
Personal reasons & 1.0000 \\
Plagiarism & 0.9901 \\
Reason not specified & 0.9647 \\
Subsumed by another publication & 0.9130 \\
Typos in manuscript & 0.9703 \\
\hline
Weighted average & 0.9594 \\
\hline\hline
\end{tabular}
\caption{Zero-shot F1 scores for various publication withdrawal categories on arXiv}
\label{tab:zero-shot-f1-scores}
\end{table}
The resulting confusion matrix (see Figure~\ref{fig:zero-shot-cm}) and per-category F1 scores (see Table~\ref{tab:zero-shot-f1-scores}) reveal that our zero-shot categorization prompt for predicting reasons for manuscript withdrawals on arXiv demonstrates strong performance across multiple categories. The per-category F1-scores range from 0.7013 to 1.0, with ``personal reasons'' and ``factual/methodological/other critical errors in manuscript'' achieving the highest scores at 1.0 and 0.9967, respectively. The surprisingly perfect classification score for ``personal reasons'' was primarily due to explicit 1st person language without any technical details, and the low F1-score for ``not novel'' is likely due to the model confusing the category with ``Subsumed by another publication'' as both categories refer to another publication, but for different reasons. The weighted average F1-score of 0.96 across all categories, indicate robust overall performance of the prompt in categorizing withdrawal reasons. These results suggest that our approach successfully captures the nuances of various withdrawal categories, with particular strength in identifying critical errors, incomplete work, and personal reasons. Reliably identifying these categories has important consequences for scientific process automation and ethical creation of datasets. We will cover these in detail in sections~\ref{sec:enabling-scify} and~\ref{sec:ethical} respectively.

\section{Discussion of Retraction Categories}
\label{sec:retraction-categories}
In this section we explain, with examples, what each of the ten retraction categories entails.

\subsubsection*{Factual/methodological/other critical errors in manuscript}
This category encompasses retractions due to significant mistakes in the research process or results. These errors could range from flawed experimental designs to incorrect data analysis to proof/lemma errors, potentially invalidating the study's conclusions.
\begin{mdframed}
\texttt{The paper is withdrawn due to a fatal mistake on pp. 7. The author is now satisfied to see that the integral of (14) actually converges in the limit x -> 1/2, as opposed to the claim of the paper in pp. 7}
\end{mdframed}
\subsubsection*{Incomplete exposition or more work in progress}
Authors may retract articles that they deem incomplete or requiring substantial additional work. This often occurs when researchers realize their initial submission was premature and requires further development or refinement.
\begin{mdframed}
 \texttt{Withdrawn due to the fact that it the proposed approach is restricted to discrete chiral symmetry and not easily generalizable to continuous chiral symmetry}
\end{mdframed}
\subsubsection*{Typos in manuscript}
While seemingly minor, typographical errors can sometimes necessitate retraction, especially if they alter the meaning of critical information or data in the paper.
\begin{mdframed}
\texttt{This paper has been withdrawn because} $\mathbb{R}\%_{+}^{n}$ \texttt{should be} $\mathbb{R}_{+}^{n}$. (sic)
\end{mdframed}
\subsubsection*{Self-identified as ``not novel''}
Researchers may withdraw their work upon realizing that their findings or ideas have already been published or are not as original as initially thought, preserving the integrity of scientific contribution.
\begin{mdframed}
\texttt{This paper has been withdrawn by the author because it is a corollary of a well-known result by Monsky}
\end{mdframed}
\subsubsection*{Administrative or legal issues}
This category includes retractions due to various non-scientific reasons, such as authorship disputes, copyright infringements, or institutional policy violations.
\begin{mdframed}
\texttt{The paper has been withdrawn waiting for the authorization from APS to reproduce two pictures published in Phys.Rev.B 63,045202 (2001)}
\end{mdframed}
\subsubsection*{ArXiv policy violation}
Articles that do not adhere to arXiv's submission guidelines or ethical standards may be retracted to maintain the platform's integrity and quality control.
\begin{mdframed}
\texttt{arXiv admin note: This submission has been removed by arXiv administrators due to unprofessional personal attack}
\end{mdframed}
\subsubsection*{Subsumed by another publication}
Authors might retract an arXiv preprint when the work is included in another preprint or publication to prevent self-plagiarism or any potential salami-slicing allegations\footnote{\url{https://en.wikipedia.org/wiki/Salami_slicing_tactics\#Salami_slicing_in_scientific_publishing}}.
\begin{mdframed}
\texttt{Most of the (correct) portion of this paper has been incorporated into the paper ``On the Markoff equation'' (arXiv:1208.4032)}
\end{mdframed}
\subsubsection*{Plagiarism}
Retractions in this category involve cases where authors have copied significant portions of others' work without proper attribution, a serious breach of academic ethics.
\begin{mdframed}
\texttt{withdrawn by arXiv administrators due to excessive unattributed and verbatim text overlap with the pre-existing Wikipedia article on redshift}
\end{mdframed}
\subsubsection*{Personal reasons}
Sometimes, authors may choose to withdraw their work for personal circumstances unrelated to the quality or content of the research itself.
\begin{mdframed}
\texttt{This version was posted without enough prior discussion with my collaborator. My collaborator would prefer it not to be posted at this time}
\end{mdframed}
\subsubsection*{Reason not specified}
This category includes retractions where authors or arXiv administrators have not provided a clear explanation for the withdrawal, potentially due to privacy concerns or other undisclosed factors.
\begin{mdframed}
\texttt{The authors have decided to withdraw this submission. Clarifications/corrections, if any, may follow at a later date}
\end{mdframed}

\section{Insights from Retraction Categorization}
\label{sec:insights-from-cat}
\begin{figure}
    \centering
    \includegraphics[width=1\linewidth]{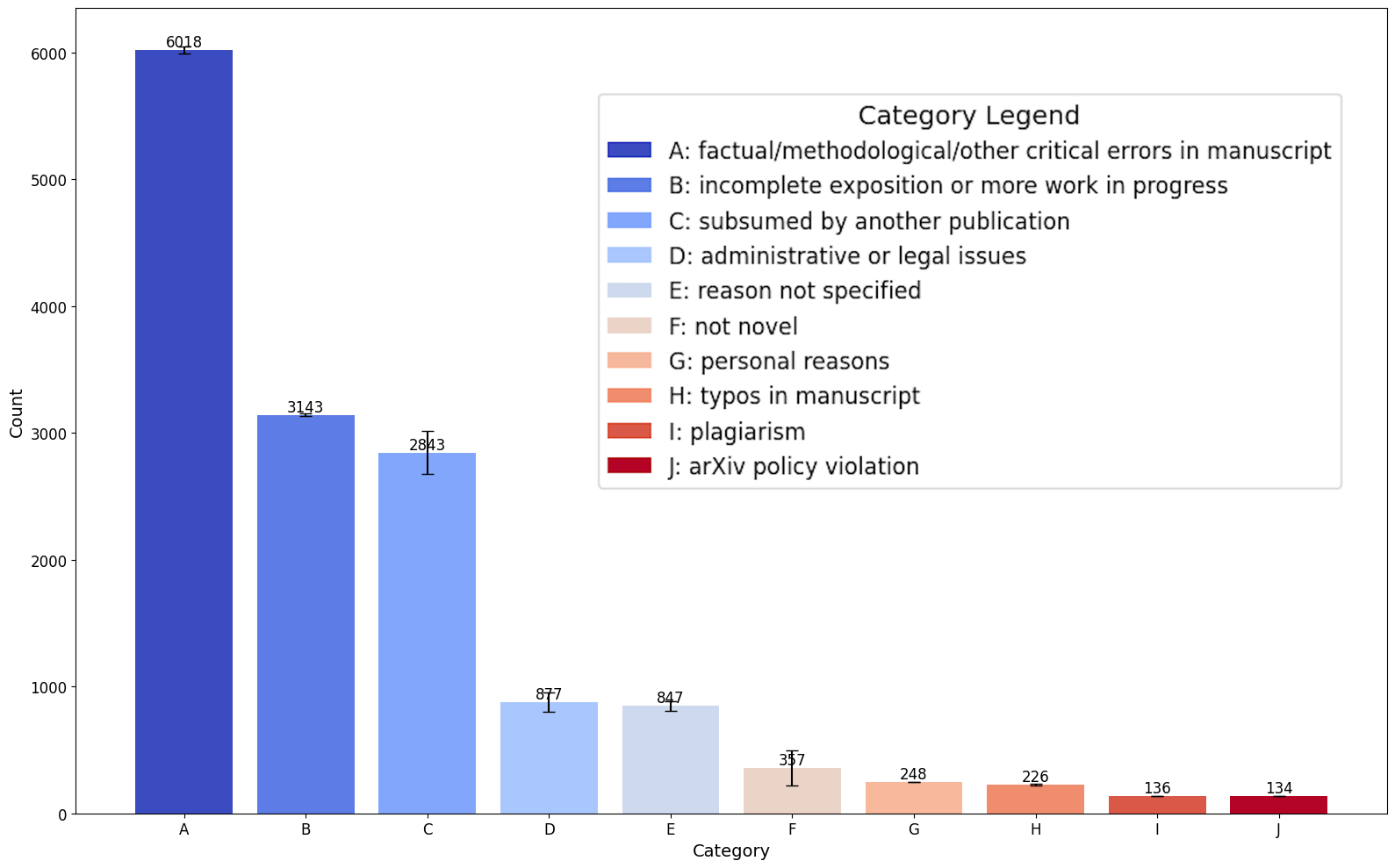}
    \caption{Distribution of reasons for paper withdrawals on arXiv. The histogram shows the frequency of different withdrawal categories, ranging from critical errors to policy violations. Each category is represented by a letter (A-J) and color-coded for clarity. Error bars are derived from categorization error rates computed via human evaluation (c.f. Section~\ref{sec:zero-shot-comment-cat}). For insights from this chart, see Section~\ref{sec:insights-from-cat}.}
    \label{fig:categories-chart}
\end{figure}
Figure~\ref{fig:categories-chart} illustrates the distribution of reasons for paper withdrawals on arXiv. Notably, the most common reason, accounting for 6,018 cases (approximately 40\% of the classifications), is "factual/methodological/other critical errors in manuscript" (category A). This is followed by "incomplete exposition or more work in progress" (3,143 cases, category B) and papers "subsumed by another publication" (2,843 cases, category C). Less frequent reasons include administrative or legal issues, unspecified reasons, and self-assessed lack of novelty. Interestingly, issues such as plagiarism (136 cases) and arXiv policy violations (134 cases) occur surprisingly rarely on the arXiv platform. This is in contrast to retraction studies on journals where plagiarism is one of the top reasons for retractions~\cite{Katavi2014}. We hypothesize this due to the nature of the pre-prints and the arXiv platform itself, where the emphasis is on sharing breaking work and the platform's automated mechanisms for plagiarism detection might deter folks from submitting plagiarized content in the first place.

\begin{figure}[h]
    \centering
    \includegraphics[width=1\linewidth]{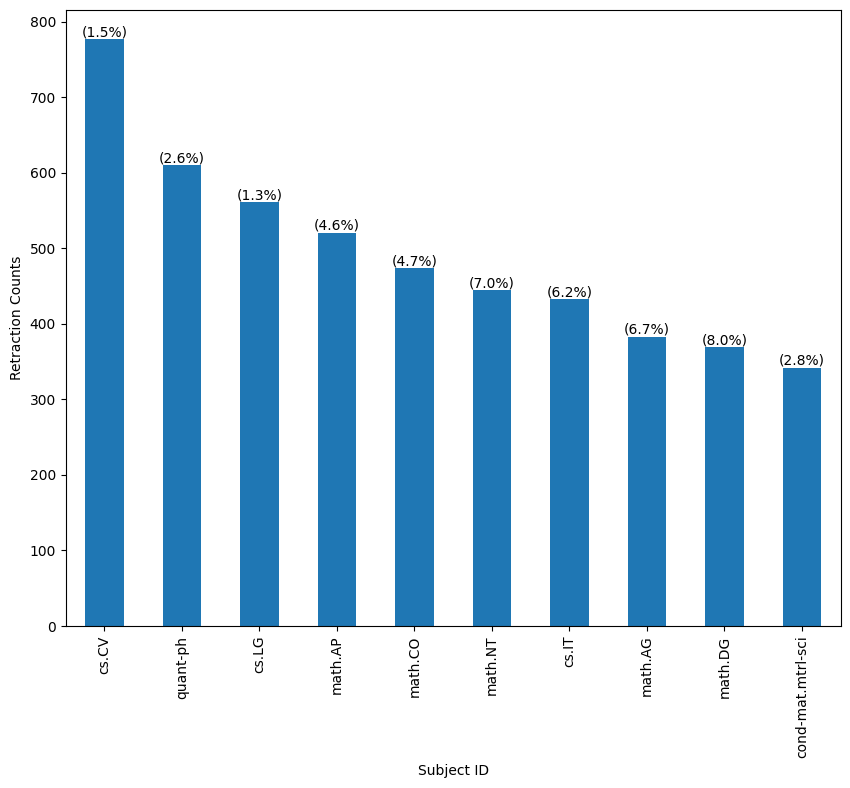}
    \caption{Top 10 arXiv subject categories with their retraction counts. AI topics, such as Computer Vision and Machine Learning (CS.LG), and Quantum Physics occupy the top, with Materials Science at the 10th place. When a preprint is cross-listed in multiple categories, we count it in each applicable category. The annotations in parentheses show retraction rates as percentages for each category. }
    \label{fig:withdraws-by-subject}
\end{figure}

If we look at retraction counts by subject category, Figure~\ref{fig:withdraws-by-subject}, the top three categories are in AI (cs.CV, cs.LG) and Quantum Physics, and Materials Science (cond-mat.mtrl-sci) is at the 10th place. However, cs.CV, cs.LG, and cs.CL are currently the largest submission categories on arXiv, so the absolute counts do not give a full picture. The retraction rates, shown in parentheses, reveal a different pattern: math.CO and math.DG show notably higher retraction rates (4.7\% and 8.0\% respectively) compared to cs.CV and cs.LG (1.5\% and 1.3\%). This suggests that while AI-related fields experience more retractions in absolute terms, certain mathematics sub-fields face higher relative frequencies of retraction. Additionally, the presence of multiple mathematical categories (math.AP, math.CO, math.NT, math.AG, math.DG) in the top 10 indicates systematic challenges in mathematical research validation, despite lower absolute retraction counts compared to AI fields. Interestingly, if we drill down each of these subject areas (see Figure~\ref{fig:categories_retractions_subject}) their retraction distributions seem similar to the global trends observed in Figure~\ref{fig:categories-chart}.

\section{Enabling SCIentific FeasibilitY (SciFy) Studies}
\label{sec:enabling-scify}
We want to highlight \DatasetName-\textsc{SciFy}, an enriched subset of \DatasetName~that includes scripts for parsed full text PDFs specifically designed to facilitate scientific feasibility studies. The creation of this dataset was motivated by a deep dive into the largest (greater than 40\% of our dataset) category of withdrawal reasons --- ``Factual/methodological/other critical errors in manuscript'' --- corresponding to 6,018 pre-prints. We clustered comments in this category to understand major themes, and we discovered eight themes:

\begin{enumerate}[nolistsep]
    \item \textbf{Errors in Proofs:}  Many authors point out errors specifically in the proofs of theorems, lemmas, or propositions. These errors range from small technical mistakes to fundamental flaws that invalidate the entire proof or result. Examples include phrases such as ``error in the proof of the main theorem'' or ``mistake in lemma.''
    \item \textbf{Misconceptions in Theoretical Foundations:} Some statements describe conceptual or theoretical misunderstandings, such as misinterpreting fundamental assumptions or using incorrect mathematical models. Phrases like ``misconception about the monodromy argument'' or ``crucial logic error'' fall into this category.
    \item \textbf{Issues in Experiment Design or Data Analysis:} Another significant theme involves problems related to experiment design, such as errors in data preprocessing, incorrect experimental setups, or misapplication of methodologies. Statements include ``error in experimental results'' and ``incorrect data analysis''.
    \item \textbf{Calculation and Numerical Errors:} A recurring theme is the discovery of calculation errors that led to incorrect results or conclusions. This might involve specific equations, constants, or algorithms that were miscomputed.
    \item \textbf{Gaps in Mathematical Arguments:} Many authors cite gaps in their logical or mathematical arguments that they were unable to resolve, thus rendering the paper incomplete or incorrect. Statements like ``gap in the proof'' or ``unfixable flaw'' are common for this subcategory. Note that this is categorically different from \#1 "Errors in proofs", as these gaps may exist outside of the proof body.
    \item \textbf{Flawed Methodologies:} Several authors mention errors in the methodology section, often resulting from flawed approaches or the need for major revisions to the proposed methods. For instance, ``error in the methodology section'' or ``method limitations for the application''.
    \item \textbf{Incorrect Assumptions or Misinterpretations:} Authors also highlight incorrect assumptions or misinterpretations that affect the validity of their results, often leading to the paper being retracted or withdrawn. This includes statements like ``misreading of the primary source'' or ``incorrect assumptions in the model''.
    \item \textbf{Errors in Figures or Visual Data:} A subset of statements refers to errors in figures, charts, or visual representations that mislead or contradict the conclusions of the paper. Examples include ``wrong figures'' or ``error in illustration''.
\end{enumerate}

\noindent We hope this dataset enables research in areas such as scientific claim verification, mathematical theorem proving, and detection of discrepancies between figures/tables and text (e.g.,\citealp{wadden-etal-2020-fact};\citealp{wadden-etal-2022-scifact};\citealp{Yang2023LeanDojoTP};\citealp{Xin2023LEGOProverNT};\citealp{Song2024TowardsLL}).

\section{Ethical Considerations and Dataset Release}
\label{sec:ethical}
While authors who retract flawed works help maintain scientific integrity, retractions remain a sensitive topic that can make some authors uncomfortable. Furthermore, retractions categorized as ``personal reasons'' divulge sensitive and sometimes potentially embarrassing information from the authors. For example, one author wrote ``I am ashamed to have written this paper'' (sic) as their retraction comment. While this information is all public and under Creative Commons, an aggregated version of the withdrawal comments makes it easy for anyone to spot such information at scale. Caution must be exercised in handling and disseminating aggregated data.

That said, we also want to encourage open research, including replication of this work. Towards that end, we are releasing this \DatasetName and \DatasetName-\textsc{SciFy} while taking several measures to protect arXiv authors from any potential embarrassment and give them control over their data. We do so with the following four concrete steps:
\begin{enumerate}
\item We exclude rows in the data categorized as ``personal reasons''. This protects authors who have divulged potentially sensitive or embarrassing information.

\item We scrub all PII from extracted retraction comments as detailed in~\ref{sec:comment-extraction}.

\item To limit distribution on a need basis, we will be releasing this dataset via HuggingFace's ``gated access'' program~\cite{huggingfaceGatedDatasets}.

\item Finally, to provide authors sovereignty over their data, we will also be working with HuggingFace to institute a ``right to be forgotten''~\cite{zhang2024rightforgotteneralarge} policy, where authors can request a specific arXiv ID to be excluded from the released dataset.
\end{enumerate}

The near-perfect F1-scores for ``personal reasons'' detection (see Table~\ref{tab:zero-shot-f1-scores}) to filter such comments along with the other measures listed above make us comfortable in releasing this data responsibly.

\section{Related Work}
\label{sec:related-work}
Our work intersects with research areas in scientific literature analysis, retraction studies, and dataset creation for scientific integrity. The majority of systematic retraction studies have focused on biomedical sciences. Wang et al.~\citeyear{Wang2017} conducted a comprehensive analysis of retractions in neurosurgical publications, finding that misconduct accounts for a significant portion of retractions. Similar studies in orthopedics~\cite{Call2024ASR} and spine surgery~\cite{Levett2023PublicationRI} have highlighted the importance of understanding retraction patterns in specific disciplines. However, systematic studies of retractions in computer science, particularly in preprint repositories, have been notably absent until now. Our work fills this gap by providing the first comprehensive analysis of withdrawals on arXiv.

    Recent years have seen growing interest in datasets supporting scientific integrity research. Wadden et al. (2020, 2022) introduced SciFact and SciFact-open for scientific claim verification, while~\cite{Agarwal2024LitLLMAT} developed tools for systematic literature review. These efforts primarily focus on published papers rather than withdrawn ones. \DatasetName{} complements these datasets by providing examples of work that authors themselves identified as problematic, offering valuable training data for automated scientific verification systems. The emergence of AI-driven science (\citealp{lu2024aiscientistfullyautomated};~\citealp{Agarwal2024LitLLMAT}) has increased interest in automated assessment of scientific work. Yang et al. (2023) and Xin et al. (2023) explored automated theorem proving, while Kasanishi et al. (2023) developed methods for automated literature review. Our \DatasetName-\textsc{SciFy} dataset provides these systems with real-world examples of scientific errors and methodological flaws, potentially improving their ability to detect problematic research before publication.

Our approach to dataset release builds on recent work in responsible data sharing. Zhang et al.~(\citeyear{zhang2024rightforgotteneralarge}) discussed the implications of the "right to be forgotten" in the era of large language models, which influenced our data release strategy. We extend these principles to scientific documentation by implementing privacy protections and author control mechanisms, similar to~\cite{laurençon2023bigsciencerootscorpus16tb} and~\cite{touvron2023llama2openfoundation}.

\section{Conclusion \& Future Work}
In this work, we have presented \DatasetName, the first comprehensive dataset and analysis of withdrawn papers from arXiv. Our contributions include:
\begin{itemize}
\item Creation and release of \DatasetName, containing over 14,000 withdrawn papers and their associated retraction comments spanning arXiv's entire history through September 2024.
\item Development of a robust taxonomy of retraction reasons, identifying 10 distinct categories that provide valuable insights into why researchers withdraw their work.
\item Demonstration of effective zero-shot categorization of retraction reasons using large language models, achieving a weighted average F1-score of 0.9594.
\item Release of \DatasetName-\textsc{SciFy}, an enriched version including parsed full-text PDFs, specifically designed to enable research in scientific feasibility studies.
\item Implementation of responsible data release practices that protect author privacy while maintaining dataset utility.
\end{itemize}

Our analysis reveals distinct patterns in arXiv withdrawals that differ significantly from those observed in traditional journal retractions. Unlike biomedical fields where plagiarism often leads withdrawals, most arXiv retractions stem from factual or methodological errors (37\%) and incomplete work (19\%). This difference highlights the unique role of preprint servers in the scientific process and suggests different quality control needs for different publication venues.

Future work could extend this research in several directions:
\begin{itemize}
\item \textbf{Cross-Platform Analysis:} Expanding the study to other preprint servers such as bioRxiv, medRxiv, and chemRxiv would enable comparative analysis of withdrawal patterns across different scientific disciplines.
\item \textbf{Temporal Analysis:} Investigating how withdrawal patterns have evolved over time could reveal trends in scientific quality control and highlight scientific disciplines or topics requiring additional attention.
\item \textbf{Enhanced Automated Verification:} Learning from patterns identified in \DatasetName-\textsc{SciFy} could lead to development of automated systems that can identify potential technical issues in drafts before submission.
\item \textbf{Citation Impact Analysis:} Studying citation patterns before and after withdrawal, could lead to better understanding of the impact of withdrawn papers on subsequent research.
\end{itemize}

These extensions would further contribute to our understanding of scientific quality control and help develop more robust systems for maintaining scientific integrity in the era of rapid electronic publishing.

\section*{Acknowledgments}
Delip and Chris would like to acknowledge Defense Advanced Research Projects Agency's (DARPA) SciFy program (Agreement No. HR00112520300) for funding this research. Jonathan's work was supported in part by the National Science Foundation under Award No. OAC-2311521 and by NASA under award No. 20-OSTFL20-0053. Any opinions, findings and conclusions or recommendations expressed in this material are those of the author(s) and do not necessarily reflect the official policy, position, or views of the National Science Foundation, NASA, the Department of Defense, or the U.S. Government.

\bibliography{acl_latex}
\section*{Appendix}
\appendix
\section{Retraction categories for four select subjects}
\label{app:categories_retractions_subject}
\begin{figure*}[htbp]
    \centering
    \includegraphics[width=1\linewidth]{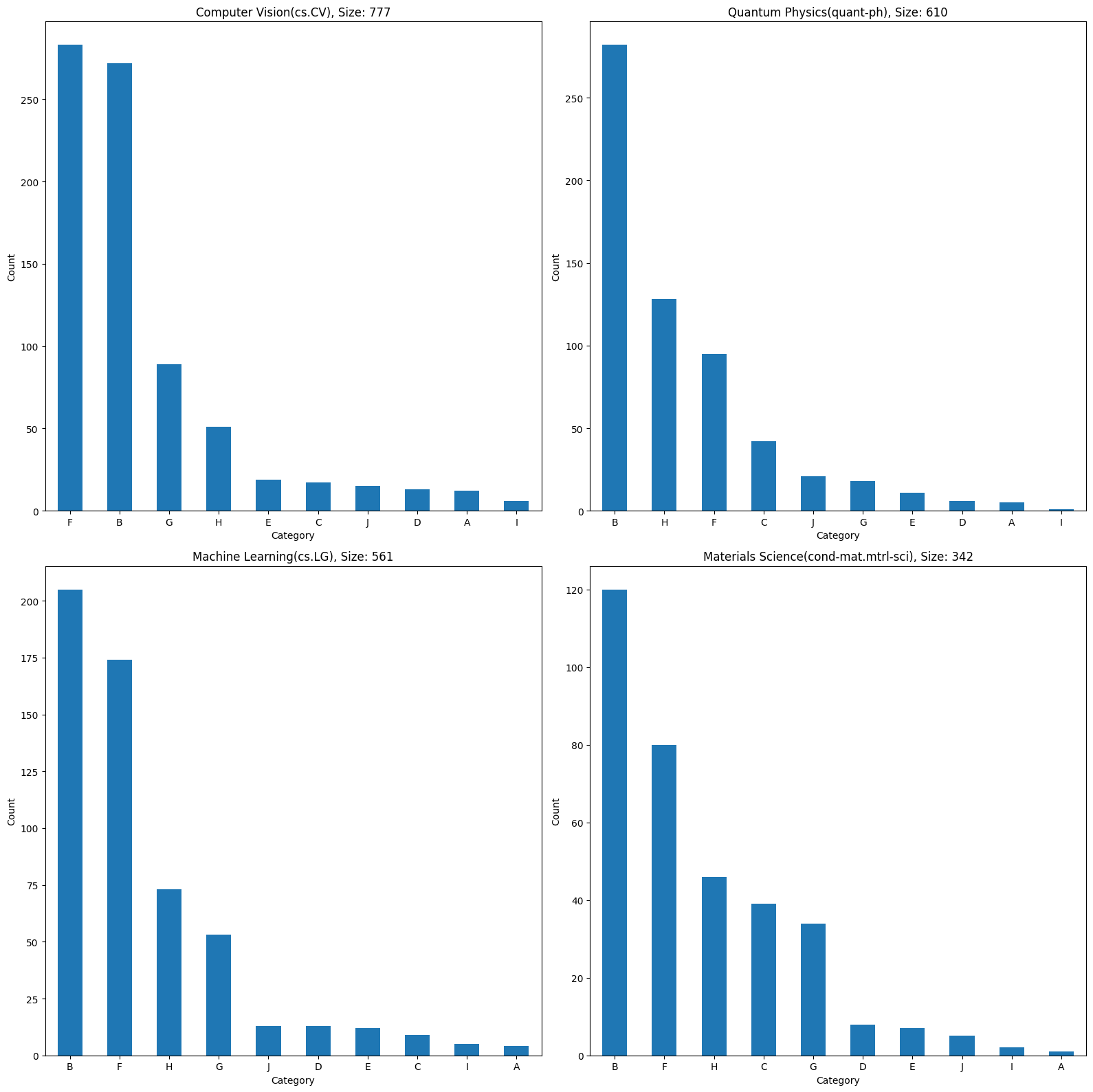}
    \caption{Retraction categories for four select
subjects -- Computer Vision, Quantum Physics/Computing, Natural Language Processing, and Materials Science (left-to-right, top-to-bottom). Legend: {\textbf{A}: `factual/methodological/other critical errors in manuscript',
 \textbf{B}: `subsumed by another publication',
 \textbf{C}: `reason not specified',
 \textbf{D}: `typos in manuscript',
 \textbf{E}: `personal reasons',
 \textbf{F}: `administrative or legal issues',
 \textbf{G}: `incomplete exposition or more work in progress',
 \textbf{H}: `plagiarism',
 \textbf{I}: `not novel',
 \textbf{J}: `arXiv policy violation'}}
    \label{fig:categories_retractions_subject}
\end{figure*}

\end{document}